\documentclass[letterpaper, 10 pt, journal, twoside]{ieeetran}  
  
\usepackage[utf8]{inputenc}
\usepackage[T1]{fontenc}
\usepackage{graphicx} 
\usepackage{subfig}
\usepackage{gensymb}
\usepackage{color}
\usepackage{cite}
\usepackage{multirow}
\usepackage{tabularx}
\usepackage{makecell}
\usepackage{multicol}
\usepackage{amsmath}

\newcommand{\change}[1]{{\color{black} #1}}
\title{UAV Localization Using Autoencoded Satellite Images}

\author{Mollie Bianchi and Timothy D. Barfoot%
    \thanks{Manuscript received: October, 15, 2020; Revised: January, 8, 2021; Accepted: February, 4, 2021. This paper was recommended for publication by Editor Sven Behnke upon evaluation of the Associate Editor and Reviewers' comments. }
    \thanks{The authors are affiliated with the University of Toronto Institute for Aerospace Studies (UTIAS):
    {\tt\footnotesize mollie.bianchi@robotics.utias.utoronto.ca}, {\tt\footnotesize tim.barfoot@utoronto.ca}} 
}

\begin{document}

\maketitle

\bibliographystyle{IEEEtran}

\markboth{IEEE Robotics and Automation Letters. Preprint Version. Accepted February, 2021}{Bianchi \MakeLowercase{\textit{et al.}}: UAV Localization Using Autoencoded Satellite Images}  

\begin{abstract}
We propose and demonstrate a fast, robust method for using satellite images to localize an Unmanned Aerial Vehicle (UAV). Previous work using satellite images has large storage and computation costs and is unable to run in real time. In this work, we collect Google Earth (GE) images for a desired flight path offline and an autoencoder is trained to compress these images to a low-dimensional vector representation while retaining the key features. This trained autoencoder is used to compress a real UAV image, which is then compared to the  precollected, nearby, autoencoded GE images using an inner-product kernel. This results in a distribution of weights over the corresponding GE image poses and is used to generate a single localization and associated covariance to represent uncertainty. Our localization is computed in 1\% of the time of the current standard and is able to achieve a comparable RMSE of less than 3m in our experiments, where we robustly matched UAV images from six runs spanning the lighting conditions of a single day to the same map of satellite images.
\end{abstract}


\section{Introduction}
\IEEEPARstart{U}{nmanned} Aerial Vehicles (UAVs) are being used for more and more applications while still remaining largely reliant on GPS. The disadvantage of a GPS-based localization system is that it is susceptible to dropout, jamming, and interference. In GPS-denied environments, the primary sensor becomes vision due to its low weight and fast computation. 

\begin{figure}
    \centering
    \includegraphics[width=.5\textwidth]{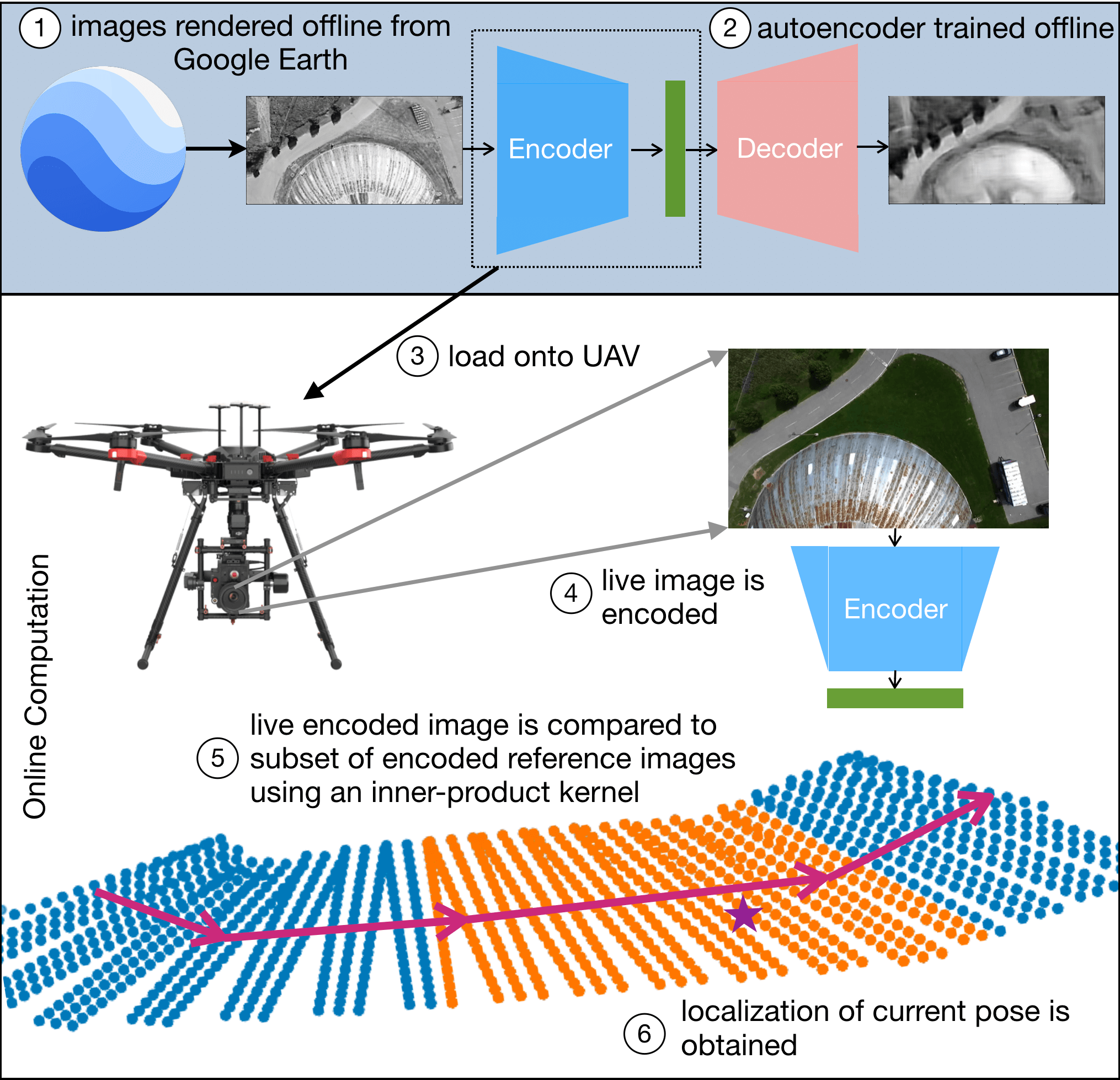}
    \caption{1. Offline before flight, images are rendered in a grid pattern around the desired path using Google Earth \cite{GoogleEarth}. An example of this grid pattern is shown at the bottom of the figure. 2.  These images are used to train an autoencoder using photometric loss and skip losses. 3. All the encoded training images and the encoder are transferred onto the UAV. 4. The live image captured by the UAV is passed through the trained encoder. 5. The  encoded live image is compared with a subset of the  encoded reference images using an inner-product kernel outputting a weight for each reference image. 6. These weights are used to compute the localization and covariance. 
    }
    \label{fig:overview}
\end{figure}

Visual Odometry (VO) is commonly used on UAVs but requires corrections to prevent drift. Visual Simultaneous Localization and Mapping (SLAM) is one solution to this issue, but its use on UAVs has primarily been demonstrated in indoor environments or small areas \cite{Bloesch2010, Shen2015, Weiss2013}. One method \cite{Warren2019} for long distance autonomous, outdoor, aerial navigation in GPS-denied environments uses the Visual Teach and Repeat (VT\&R) method \cite{Furgale2010}. By generating a locally consistent visual map on an outbound pass under manual or GPS control, the UAV is then able to return autonomously along that path without GPS. 

VT\&R is limited in that it requires a manual outbound pass and, because it relies primarily on point-feature matching (e.g., Speeded-Up Robust Features (SURF) \cite{Bay2008}), the return pass must be completed shortly after the outbound pass so that the lighting conditions along the path have not significantly changed. It does not allow for the repeated traversal of the path using a map generated much earlier. 

A unique opportunity for aerial vehicles is that there is an existing database of satellite images covering the entire world in Google Earth (GE). In many areas, these satellite images have been used to generate a detailed 3D reconstruction of a scene from which it is possible to render an image at any desired pose. In \cite{Patel2020}, the idea was proposed to replace the manual outbound pass in VT\&R with a virtual pass in GE. 
 \begin{figure*}
     \centering
     \includegraphics[width=0.97\textwidth]{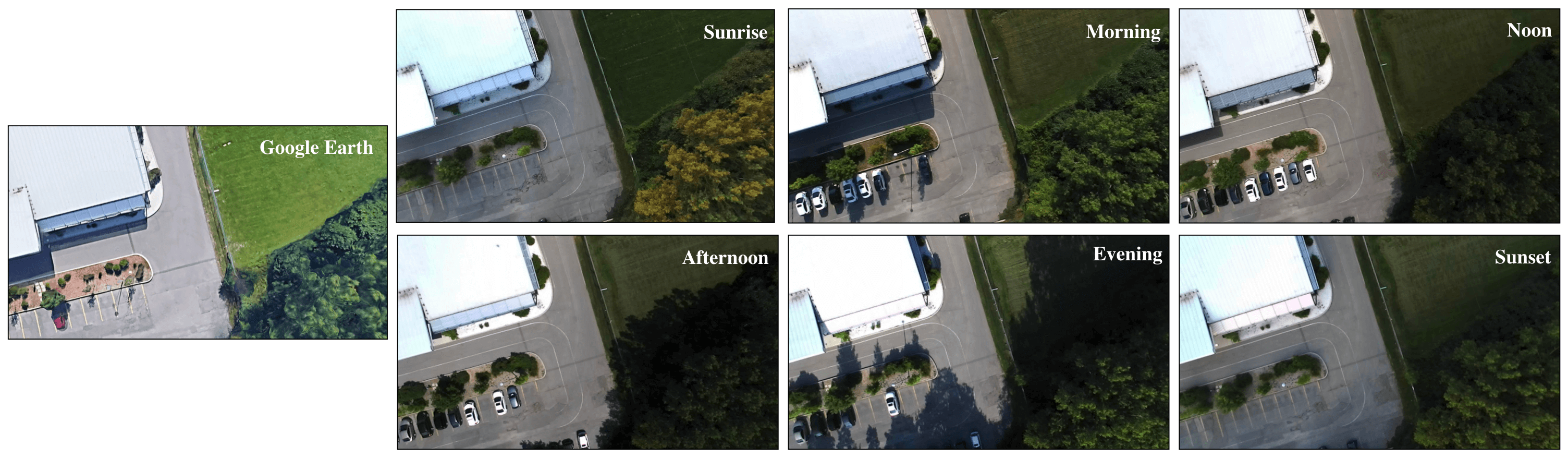}
     \caption{An image from each of the six lighting conditions in the dataset and a corresponding image rendered from GE \cite{GoogleEarth} is shown. The shadows present in the GE images most closely resemble those present in the morning lighting condition. The shadows in the afternoon and evening datasets appear on the opposite side of objects as compared to GE.  }
     \label{fig:lc_imgs}
 \end{figure*}

The largest challenge with this idea is finding a way to accurately and robustly localize real live images captured from a UAV with the artificial images rendered from GE. Since the satellite images used for the reconstruction were captured years ago, there are differences with the live images in terms of lighting, small object movement (e.g. vehicles, trailers), large structural changes (e.g. building additions/demolitions), and unusual object reconstruction, particularly for non-rectangular based objects like trees. This makes it difficult for feature-based methods to obtain accurate and robust results in many cases.  

Patel et al. \cite{Patel2020} used multiple GE images rendered around the desired flight path and used mutual information (MI) to search for the best alignment with the live image. This approach was computationally expensive and would require storing thousands of full-sized images on board the UAV. It is not capable of running in  real time. 

This work introduces a new method to use prerendered satellite images that is fast and storage efficient. As in \cite{Patel2020}, images are rendered around the desired flight path in GE. An autoencoder is trained on these path-specific images to compress them to a much smaller vector representation. The same autoencoder is used to compress the live images as well. The compressed live image vector is compared to all nearby compressed GE image vectors through an inner-product kernel. This results in weights associated with each of the corresponding GE image poses. From these weights, a localization for the longitude, latitude, and heading with accompanying covariance is computed. This method has been demonstrated on a real UAV dataset of images along a 1.1km path at six different times of day covering several lighting conditions. In comparison with \cite{Patel2020}, we are able to achieve the same accuracy performance on image registration and run in 1\% of the computation time.

The rest of this paper is organized as follows.  Section II reviews the related work from the literature.  Section III discusses our methodology.  Sections IV and V provide our experimental results on a real UAV dataset.  Section VI wraps up with our conclusions and suggestions for future work.

\section{Related Work}

\subsection{Aerial Visual Localization}

Early works in visual aerial localization looked at using edge detection \cite{Conte2008} or a combination of classical techniques with learned semantic segmentation \cite{Nassar2018}. Both these approaches perform better at high-altitude flights where more structure is present in the images and suffer in areas comprising mainly grass and trees. There have been more recent feature-based methods that match street view images to images from a ground robot \cite{Agarwal2015} and a UAV \cite{Majdik2013, Majdik2015}. Place recognition is performed using a visual bag-of-words technique and then followed by image registration using Scale Invariant Feature Transform (SIFT) \cite{Lowe2004} keypoints. However, feature matching has been shown to contain significant numbers of outliers due to large changes in appearance and viewpoint.

The current best method for localization using satellite images is the MI based approach presented in \cite{Patel2020}. This was largely inspired by \cite{Pascoe2015, Pascoe2015_2} in which MI had been used to localize monocular camera images within a textured 3D model of the environment. Adapting this idea to a UAV, Patel et al. \cite{Patel2020} were able to achieve less than 3m and 3\degree \: Root Mean Square Error (RMSE) on low-altitude flights at six different times of day. In their work, images were rendered from GE beforehand every 3m along the path and around the path at intervals of 6m. The Normalized Information Distance (NID), which is a MI based metric, was computed between the live image and all images within 4m of the prior pose estimate (e.g., from filtering) to select the best-matching image. The alignment between this geo-referenced image and the live image was then computed by a series of coarse and refined optimizations of the warping parameters. Each step of the optimization required numerically computing the Jacobian of the NID with respect to the warping parameters.  This process was quite slow making this method incapable of running in real time. As well, the images were stored in their full $560 \times 315$ resolution resulting in large storage costs for longer paths. We build on the idea of prerendering images around a desired path in GE, but improve upon \cite{Patel2020} by eliminating the costly optimization step, improving runtime, and decreasing storage requirements.

\subsection{Autoencoders}
One of the core limitations with \cite{Patel2020} is that the images are large, each is 176,400 pixels making the MI computation slower.  A common learning-based method for compressing images is autoencoders \cite{Kingma2014}. One neural network acts as the encoder, compressing images down to some low-dimensional bottleneck. A second network upsamples this bottleneck vector back to an image with the same size as the original image. Minimizing some loss function between the original image and the recreated image, the network can learn to retain the key features in the bottleneck. There has been lots of work in this area, including new loss functions \cite{Hou2017, Ridgeway2015}, adversarial autoencoders \cite{Makhzani2015}, and combining autoencoders with neural autoregressive models \cite{Chen2016}. In this work, we use an autoencoder architecture based on \cite{Hou2017} to compress our images.

\subsection{Kernels}
Kernels, such as the inner-product or exponential kernel, are often used for matching patches between images, such as in \cite{Grauman2005, Zhang2007}. They provide a measure of the similarity between the two patches, but they are not commonly used for comparisons of whole images due to the high number of pixels involved. We use kernels on the autoencoded representations of whole images. Since these autoencoded images are small enough to quickly compute a kernel between them, it eliminates the need for extracting and matching patches from an image.

\change{
\subsection{Learned Pose Estimation}
Using learned methods to directly compute the poses of objects in images, or the relative pose change between two images has been the focus of many works \cite{Melekhov2017, kendall2015posenet, tekin2018real}. However, these approaches are limited by the available training data. Real data is expensive to collect and label, and synthetic data does not typically generalize directly to the real world. Alternatively, Sundermeyer et al. \cite{Sundermeyer2018} use a similar method to what is proposed in this work to perform 6D Object Detection. Instead of explicitly learning from 3D pose annotations during training, they implicitly learn representations from rendered 3D model views. Using a denoising autoencoder, they generate a codebook containing the encoded representations of tens of thousands of rendered images of the desired object at uniformly distributed poses. The same autoencoder is used to encode a live image and a cosine similarity metric is used to match the live image with the closest poses from the codebook.}

\begin{figure}[h!]
    \centering
    \includegraphics[width=0.95\columnwidth]{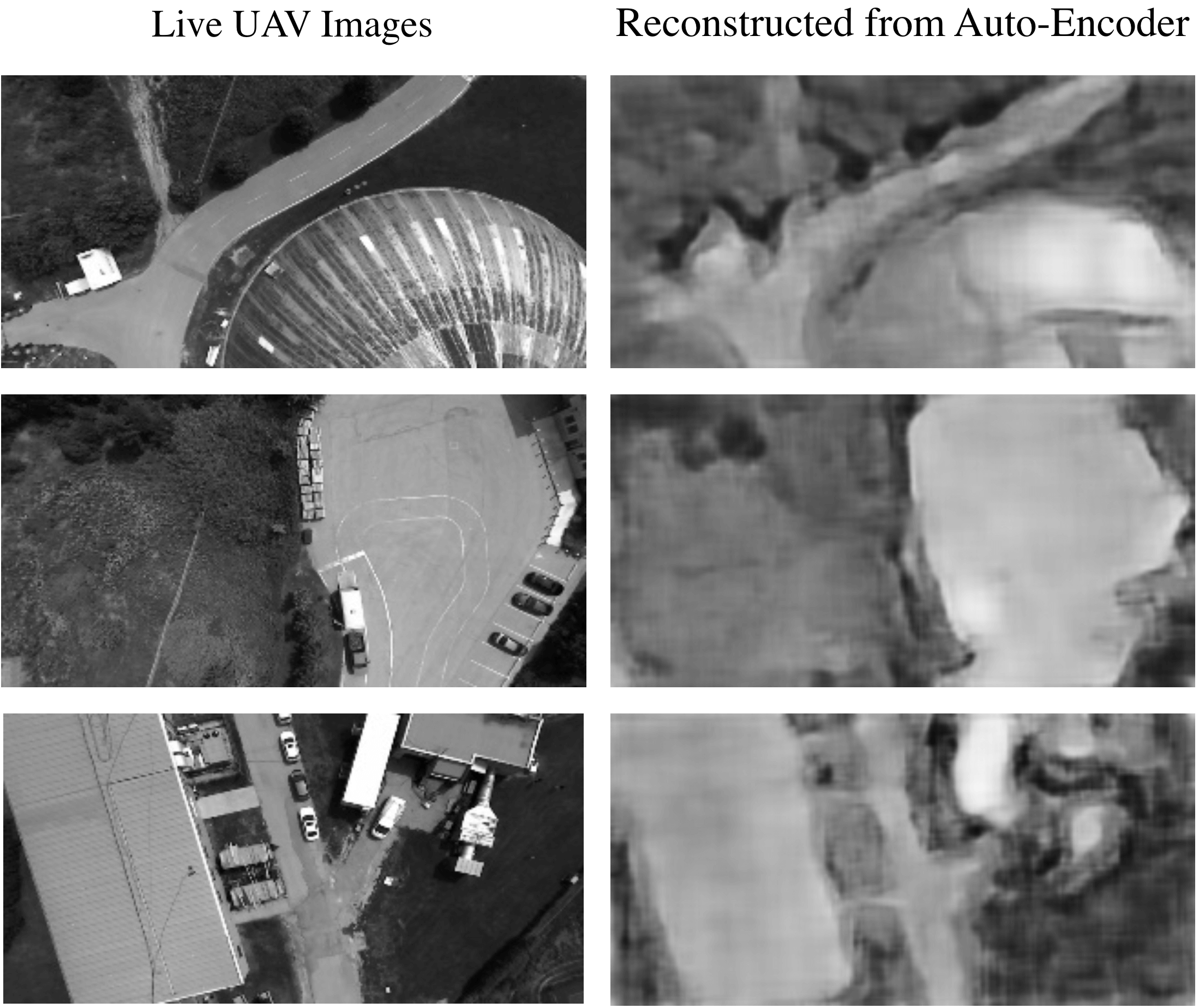}
    \caption{\change{On the left are real images collected by the UAV along the path. On the right is the corresponding image after passing through the autoencoder and decoder which were only trained on GE images. }}
    \label{fig:recon_real_images}
\end{figure}

\section{Methodology}
The proposed approach can be divided into several steps as depicted in Figure~\ref{fig:overview}. Offline, images are rendered from GE around the desired flight path. Then an autoencoder is trained for this specific path using these images. These images are encoded and saved after passing through the trained network. \change{While the offline processing is significant, it only needs to be completed once per path and would eliminate the need for manual mapping flights before each autonomous flight as in \cite{Warren2019}.} In the online component of the pipeline, weights for a subset of these autoencoded reference images are computed using an inner-product kernel computation with an autoencoded live image. The localization and corresponding covariance are then computed from these weights. Finally, outlier rejection is performed based on the covariance estimate from the previous step.
 
\subsection{Pre-Flight Image Collection}
Using a desired path, images are rendered from GE \cite{GoogleEarth} at the intended orientation every 0.5m along the path. Additional images are rendered at 0.5m lateral offsets out to 5m to either side of the path. This requires 42 images for each meter of the path. This coverage could be modified based on expected performance of the UAV. For example, if you are expecting the UAV to operate in windy conditions more images could be rendered further from the path. Regardless, this leads to a high number of images for non-trivial path lengths. Storing and comparing these images in full size would be infeasible. The next step and key aspect of this method is to use an autoencoder to compress the images to a low-dimensional representation while maintaining the key features of each image such that comparing the compressed images using a kernel yields sensible results. 

\subsection{Autoencoder}
\begin{figure}[h!]
    \centering
    \includegraphics[width=0.97\columnwidth]{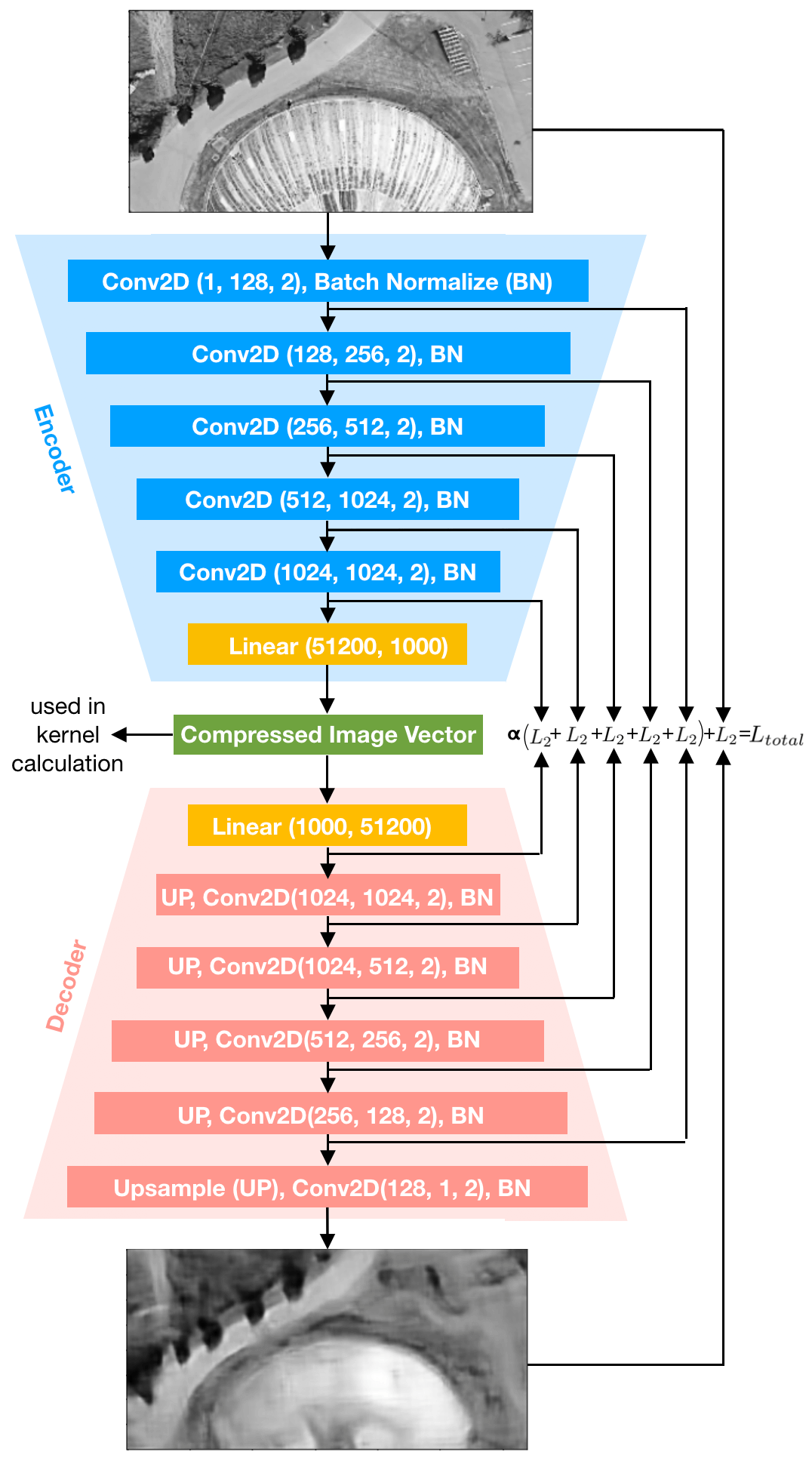}
    \caption{The autoencoder architecture is based on \cite{Hou2017}. It uses traditional photometric loss between the input image and the reconstructed image as well as skip losses between the corresponding layers in the encoder and decoder.}
    \label{fig:vae_arch}
    \vspace{2mm}
\end{figure}

With a focus on place-specific excellence, the autoencoder is trained solely using the precollected images from GE for that desired path. A new autoencoder would need to be trained for each path. The autoencoder architecture is based on \cite{Hou2017} as implemented in \cite{autoencoder_git}. The input is a $320 \times 160$ greyscale GE image. The encoder is composed of six layers. Each of the first five layers perform a 2D convolution with a stride of two followed by a batch normalization layer. The number of channels double as indicated in Figure \ref{fig:vae_arch}. Finally, a linear layer maps the output of the final convolution layer to the bottleneck vector. Different sizes were experimented with for the dimension of the bottleneck. A bottleneck of dimension 1000 was selected as it was the smallest size that could still achieve the desired accuracy. 

The decoder behaves opposite the encoder. A linear layer first maps the bottleneck variable to 1024 channels. This is then passed to the first of five layers, each of which performs upsampling by a factor of two, followed by convolution with a stride of three, and batch normalization. The number of channels is halved in each layer until an output greyscale image with the same dimension as the input image is generated. To obtain the compressed image vector, the output after only the encoder part of the network is used.

The loss function used to train the network is a combination of photometric loss between the input and output images, \change{i.e., L = $(I_{input} - I_{output})^2$}, and L2 loss between the outputs of corresponding layers referred to as skip losses. These additional skip losses \change{are weighted with a value of 0.01 and} encourage the decoder to learn the reverse behaviour of the encoder and was found to improve performance on the real image validation sets. The network was trained for 20 epochs with a learning rate of 1e-4. \change{For the path used in the experiments, the network was trained with approximately 48,000 images. On a Nvidia DGX Station using a single Tesla V100 GPU training took around 20 hours to complete. } 

It is important to stress that the network is trained only on images from GE, but it is still able to generalize to real-world images with different lighting conditions. Thus it is able to be trained before having to actually fly the path. \change{Figure \ref{fig:recon_real_images} shows some examples of real images reconstructed after passing through the autoencoder. While the reconstructions are not as sharp as the reconstruction of the training data as seen in Figure \ref{fig:vae_arch}, the main structures in each image are preserved. For our application of comparing the encoded live image to encoded reference images, the network achieves sufficient generalization performance. It may be interesting for future work to look into using data augmentation during training to further improve generalization performance similar to \cite{Sundermeyer2018}. }

\subsection{Localizing Using Kernels}
To minimize time spent loading the autoencoded GE reference images, all image vectors are stacked and loaded into a $1000 \times N$ dimensional matrix denoted ${\mathbf{Y}}_{\rm ge}$, which is loaded onto the GPU. Then, based on a prediction of the current live image pose, $\mathbf{Y}_{\rm ge}$ is indexed to include only the reference images that are 4m ahead and behind along the path, which is the same search area used in \cite{Patel2020}. The live image is passed through the trained autoencoder and the resulting compressed $1000 \times 1$ vector is denoted as $\mathbf{y}$. The weights, $\mathbf{w}$, are computed for the subset of autoencoded GE reference images using a basic inner-product kernel:
\begin{equation}\label{eq:kernel}
\mathbf{w} =  \mathbf{Y}_{\rm ge}^T \mathbf{y}.       
\end{equation}

$\mathbf{w}$ contains a similarity measurement between the live image and each of the reference images. Since there are 336 images being used for comparison, many of these images have low, but non-zero, weights. These weights tend to pull the mean value towards the centre of the area covered by the reference images. To prevent this and get a result closer to the images with the highest weights, a new set of weights, $\mathbf{w}_{\rm th}$, is created by setting all values of the weights less than one standard deviation of the max weight to zero. The thresholded weights are then normalized:
\begin{equation}
    \bar{\mathbf{w}}_{\rm th} = \frac{\mathbf{w}_{\rm th}}{\sum_i {w_{{\rm th},i}}}.
\end{equation}

The longitude and latitude coordinates of each reference image are stacked in a $2 \times N$ matrix $\mathbf{X}_{\rm ge}$. The thresholded weights are used to compute the localization for the longitude and latitude according to:
\begin{equation}\label{eq:mean}
    \hat{\mathbf{x}} = \begin{bmatrix} \hat{x} \\ \hat{y} \end{bmatrix} = \mathbf{X}_{\rm ge} \bar{\mathbf{w}}_{\rm th}.
\end{equation}
The covariance is computed using the original weight values:
\begin{equation}
    \mathbf{P} = \sum_i w_i (\mathbf{x}_{{\rm ge},i} - \hat{\mathbf{x}})(\mathbf{x}_{{\rm ge},i} - \hat{\mathbf{x}})^T.
\end{equation}
Some examples showing the localization, covariance, and weights generated for each nearby GE reference image are shown in Figure \ref{fig:testpoints}.

Instead of rendering reference images at multiple headings and including them in the previous computation, the heading computation is performed after the above step. The reference image with the largest weight is selected for comparison, $\mathbf{y}_{\rm ge}^{\star}$. The uncompressed live image is then rotated in 1\degree \: increments between -5\degree and 5\degree. All these rotated images are autoencoded and stacked into an $1000 \times 11$ matrix, $\mathbf{Y}_{\theta}$. The kernel computation from \eqref{eq:kernel} is repeated:
\begin{equation}
\mathbf{w}_{\theta} =  \mathbf{Y}_{\theta}^T \mathbf{y}_{\rm ge}^{\star} .    
\end{equation}
These weights are normalized, $\bar{\mathbf{w}}_{\theta}$, and then used to compute a heading measurement following the same procedure as in \eqref{eq:mean} using a stacked vector of the rotation values, $\mathbf{x}_{\theta}$:
\begin{equation}
\hat{\theta} =  \mathbf{x}_{\theta}^T \bar{\mathbf{w}}_{\theta} .    
\end{equation}
This mean heading value, $\hat{\theta}$, is then added to the heading of the selected reference image, $\beta$, to get a global heading, $\hat{\theta} + \beta$. While not included here, a similar process could be used to get an estimate for altitude without having to render additional reference images.  The full localization is then:
\begin{equation}
    \hat{\mathbf{p}} = \begin{bmatrix}
    \hat{x} \\
    \hat{y} \\
    \hat{\theta} + \beta
    \end{bmatrix}
\end{equation}

\change{One of our earlier approaches for localization was to use the position of the image with the highest similarity measurement. This yielded fairly similar results to the weighted average approach except that it was limited by the grid spacing of the reference images and more susceptible to outliers.}

\begin{figure}[h!]
    \centering
    \includegraphics[width=\columnwidth]{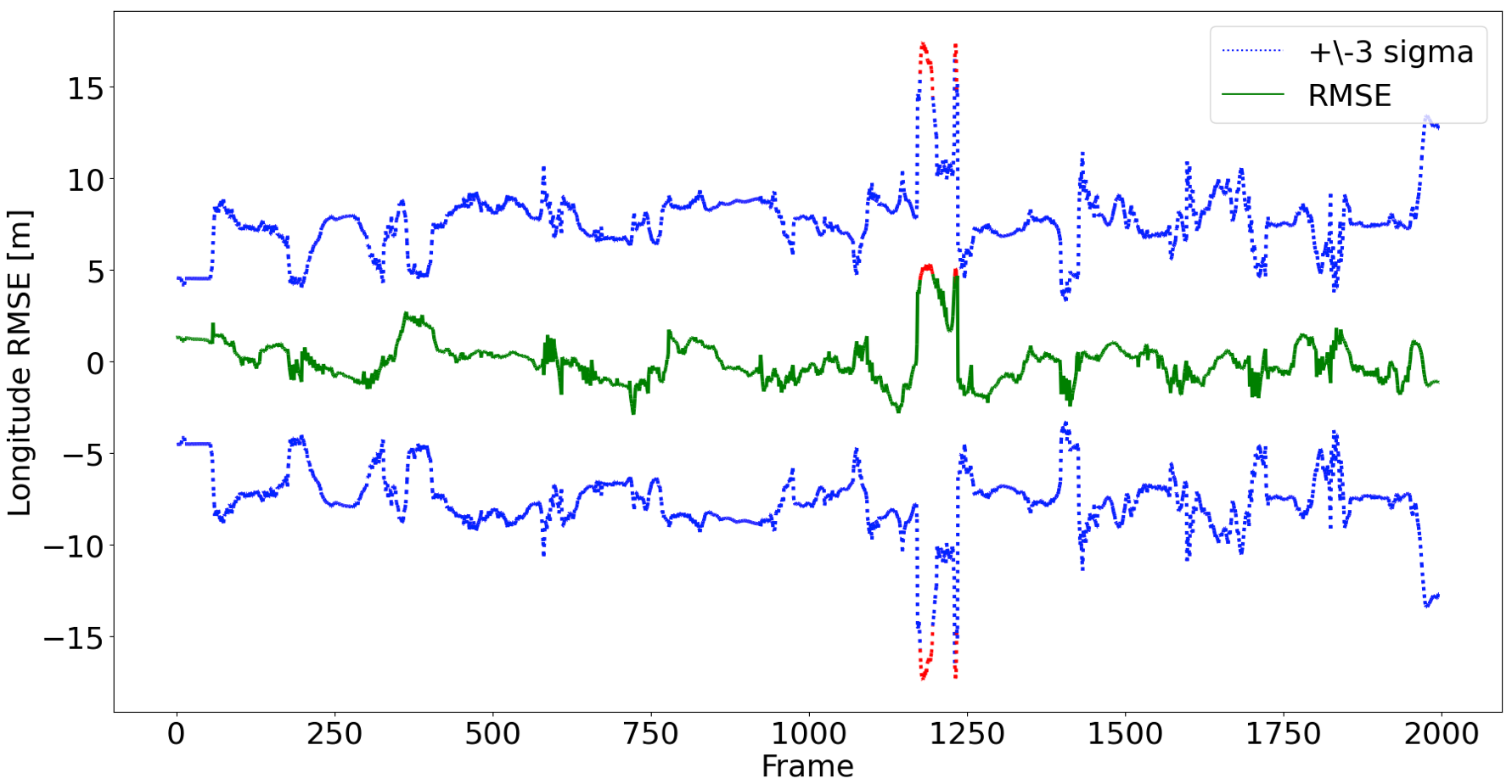}
    \caption{Plots of the RMSE for the longitude coordinates of the sunrise test are shown here with accompanying 3$\sigma$ uncertainty envelope. The registrations that were rejected due to either $\sigma_{\textrm{long}}$ or $\sigma_{\textrm{lat}} > 5$ are shown in red.}
    \label{fig:errorplots}
\end{figure}
 
\subsection{Outlier Rejection} \label{sec:outlier}
Since we compute a covariance with our localization based on the weights, we can also use this to reject outliers. When the weights have a single narrow peak away from the edges of the area covered by the reference images, the $\sigma^2_{\rm long}$ and $\sigma_{\rm lat}^2$ values are small. When the weights are more spread out with a less-well-defined peak, when there are multiple peaks, or when the peak occurs very close to the edge of the reference area, this results in larger values for $\sigma^2_{\rm long}$ and $\sigma_{\rm lat}^2$. We reject localizations that have either $\sigma_{\rm long}$ or $\sigma_{\rm lat}$ greater than 5. Figure \ref{fig:errorplots} plots the RMSE for the longitude coordinates with accompanying covariance. Rejected registrations are indicated in red. 

\begin{figure}[b]
    \centering
    \includegraphics[width = 0.9\columnwidth]{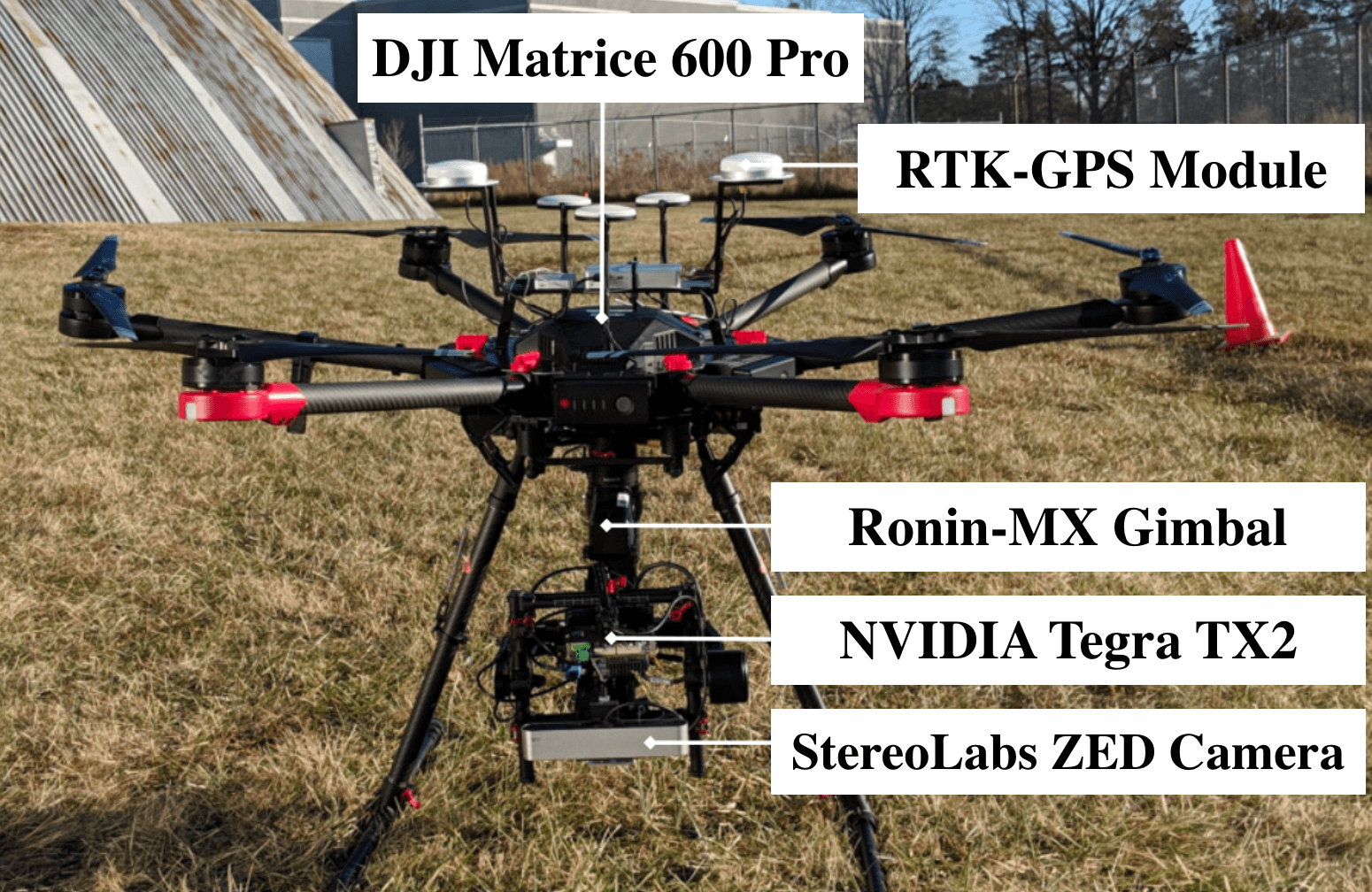}
    \caption{The dataset was collected by a 3-axis gimballed stereo camera on a multirotor UAV.}
    \label{fig:uav_setup}
\end{figure}


\begin{table*}
    \renewcommand{\arraystretch}{1.1}
    \vspace{4mm}
    \caption{Comparison of Errors for Successful Registrations}
    \begin{tabular}{l||c|c|c||c|c|c|c|c|c|c|c|c||}
         \multirow{3}{5em}{\thead{\textbf{Lighting} \\\textbf{Condition}}} &  \multicolumn{3}{c||}{\multirow{2}{4em}{\makecell{\textbf{Registration} \\\textbf{Success [\%]} \\[5pt]}}} & \multicolumn{9}{c||}{\textbf{Successful Registrations RMSE}}\\
          & \multicolumn{3}{c||}{}& \multicolumn{3}{c|}{longitude [m]} & \multicolumn{3}{c|}{latitude [m]} & \multicolumn{3}{c||}{heading [degree]}\\
         & Ours A & Ours B & MI \cite{Patel2020} & Ours A & Ours B & MI \cite{Patel2020}& Ours A & Ours B & MI \cite{Patel2020} & Ours A & Ours B & MI \cite{Patel2020}\\
         \hline
          Sunrise & 98.8 & \textbf{99.9} & 94.7 & \textbf{1.05} & 1.17 &  1.10 & 0.97 &  0.97 & \textbf{0.71} & \textbf{0.35} & \textbf{0.35} & 2.28 \\
          Morning & \textbf{100} & \textbf{100} & 95.1 & \textbf{0.90} & \textbf{0.90} & 1.02 & 0.95 & 0.95 & \textbf{0.58} & \textbf{0.31} & \textbf{0.31} & 2.57\\
          Noon & \textbf{100} & \textbf{100} & 97.8 & 1.04 & 1.04 & \textbf{0.78} & 0.87 & 0.87 & \textbf{0.61} & \textbf{0.36} & \textbf{0.36} & 1.82\\
          Afternoon & \textbf{98.0} & \textbf{98.6} & 96.0 & \textbf{1.64} & 1.67 & 1.69 & \textbf{0.88} & \textbf{0.87} & 0.92 & \textbf{0.34} & \textbf{0.34} & 1.17\\
          Evening & 90.9 & \textbf{96.0} & 81.3 & \textbf{2.16} & 2.17 & 3.03 & 1.18 & \textbf{1.14} & 1.32 & \textbf{0.42} & \textbf{0.41} & 2.49\\
          Sunset & 97.4 & \textbf{98.7} & 87.5 & \textbf{1.37} & 1.48 & 1.95 & 0.96 & \textbf{0.94} & 1.12 & \textbf{0.36} & \textbf{0.36} & 2.64\\
    \end{tabular}
    \vspace{-4mm}
    \label{table:succ_regs}
\end{table*}

\section{EXPERIMENTAL SETUP}

\subsection{Image Registration on UTIAS Dataset}
 Image registration to obtain the longitude, latitude, and heading was performed on the same dataset as in \cite{Patel2020}. We do not focus on estimating the roll, pitch, or altitude of the vehicle as those can be measured by complementary sensors to vision. The data was collected at UTIAS using a DJI Matrice 600 Pro multirotor UAV with a 3-axis DJI Ronin-MX gimbal (see Figure \ref{fig:uav_setup}). A StereoLabs ZED camera provides stereo images at 10FPS.  The RTK-GPS system and IMU provide the vehicle pose for ground truth.
 
 This dataset consists of six traversals of a 1132m path over built-up areas with roads and buildings as well as large areas of grass and trees. Each run captures the distinctive lighting condition at different times of day: sunrise, morning, noon, afternoon, evening, and sunset. These lighting conditions are shown in Figure \ref{fig:lc_imgs}. The UAV flies at an altitude of 40m with a constant heading and the camera pointed in the nadir direction.
 
 There is an unknown offset between the RTK-GPS frame and the GE frame. So 10\% of the successful image registrations are used to align the frames. These registrations are then omitted from all error calculations.

 \section{RESULTS}
 
\change{Most vision-based localization methods rely on features. Previously, Patel et al. \cite{Patel2020} evaluated the ability of SURF features to match between GE rendered images and live images on the same path used here under various lighting conditions using the VT\&R framework in \cite{Warren2019}. Using the GE images for the teach pass and the live images for the repeat, features were only capable of producing less than 7\% successes per repeat if a successful registration is defined as having greater than 30 Maximum Likelihood Estimation Sample Consensus (MLESAC) inliers. We are not currently aware of any works localizing GE images to real UAV images at a similar altitude and orientation as our flight path other than \cite{Patel2020}.} 
 
We are able to achieve comparable results with the MI-based approach from \cite{Patel2020} on the same dataset in 1\% of the computation time. In Table \ref{table:all_regs}, we present our RMSE for the longitude, latitude, and heading for all registrations on each of the six runs. We achieve lower errors as compared to the results presented in \cite{Patel2020}. \change{ We use the same search area as in \cite{Patel2020} to select our subset of reference images, which correspond to a maximum RMSE of 6.4m.}

\begin{table}[b]
\caption{Comparison of Errors for All Registrations}
    \renewcommand{\arraystretch}{1.1}
    \begin{tabular}{l||c|c|c|c|c|c||}
         \multirow{3}{5em}{\thead{\textbf{Lighting} \\\textbf{Condition}}} &  \multicolumn{6}{c||}{\textbf{All Registrations RMSE}} \\
         & \multicolumn{2}{c|}{longitude [m]} & \multicolumn{2}{c|}{latitude [m]} & \multicolumn{2}{c||}{yaw [degree]} \\
         & Ours & MI \cite{Patel2020} & Ours & MI \cite{Patel2020}& Ours & MI \cite{Patel2020} \\
          \hline
 Sunrise    &  \textbf{1.18}  & 1.87 & \textbf{0.98}  & 1.47 & \textbf{0.35} & 2.80 \\
 Morning    &  \textbf{0.90}  & 2.24 & \textbf{0.95}  & 1.39 & \textbf{0.31}  & 2.97\\
 Noon       &  \textbf{1.04}  & 1.26 & \textbf{0.87}  & 1.02 & \textbf{0.36}  & 2.70 \\
 Afternoon  &  \textbf{1.84}  & 2.14 & \textbf{0.90} & 1.57 & \textbf{0.35} & 2.63 \\
 Evening    &  \textbf{2.53}  & 4.09 & \textbf{1.19}  & 3.63 & \textbf{0.42}  & 5.25\\
 Sunset     & \textbf{1.64} & 3.03 & \textbf{0.97}  & 1.95 & \textbf{0.37}  & 3.06 \\
    \end{tabular}
    \label{table:all_regs}
\end{table}

\change{In Table \ref{table:succ_regs}, the error results from only successful registrations are compared against the errors from successful registrations in \cite{Patel2020}. For "Ours A", we use the outlier rejection scheme described in Section \ref{sec:outlier} to reject registrations with a high covariance estimate. For "Ours B", we use the outlier rejection from \cite{Patel2020} along with our localization method. In \cite{Patel2020}, registrations are deemed unsuccessful if the localization is too far from the previous estimate. There is minimal difference in the performance between "Ours A" and "Ours B". The benefit of our outlier rejection scheme is that it is based purely on the covariance of the localization result and does not require a prior estimate.} 
\begin{figure}[h!]
\centering
\subfloat[Morning]{%
  \includegraphics[width=0.88\columnwidth]{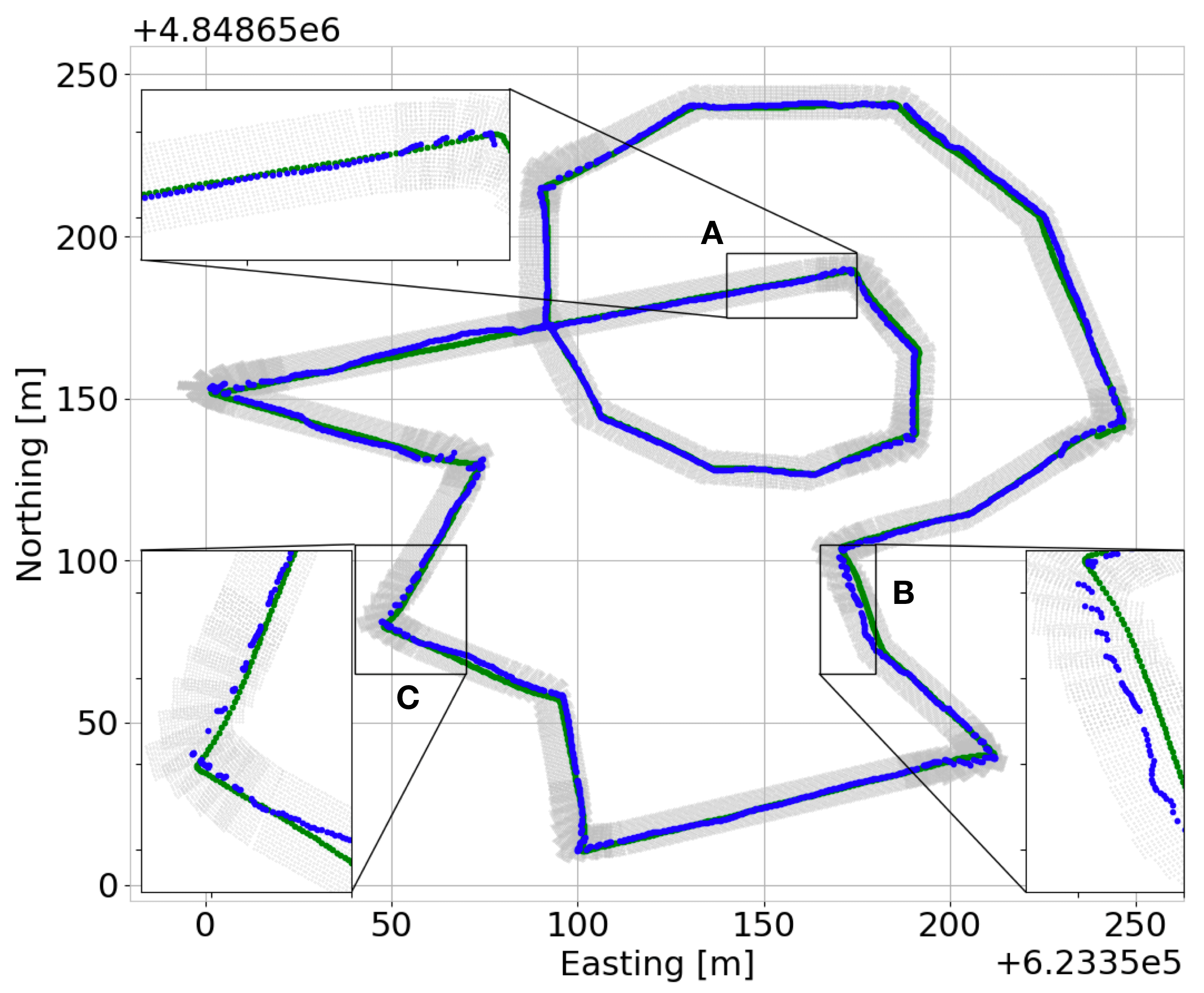}%
  \label{fig:morning}
}\\[-0.2ex]
\subfloat[Evening]{%
  \includegraphics[width=0.88\columnwidth]{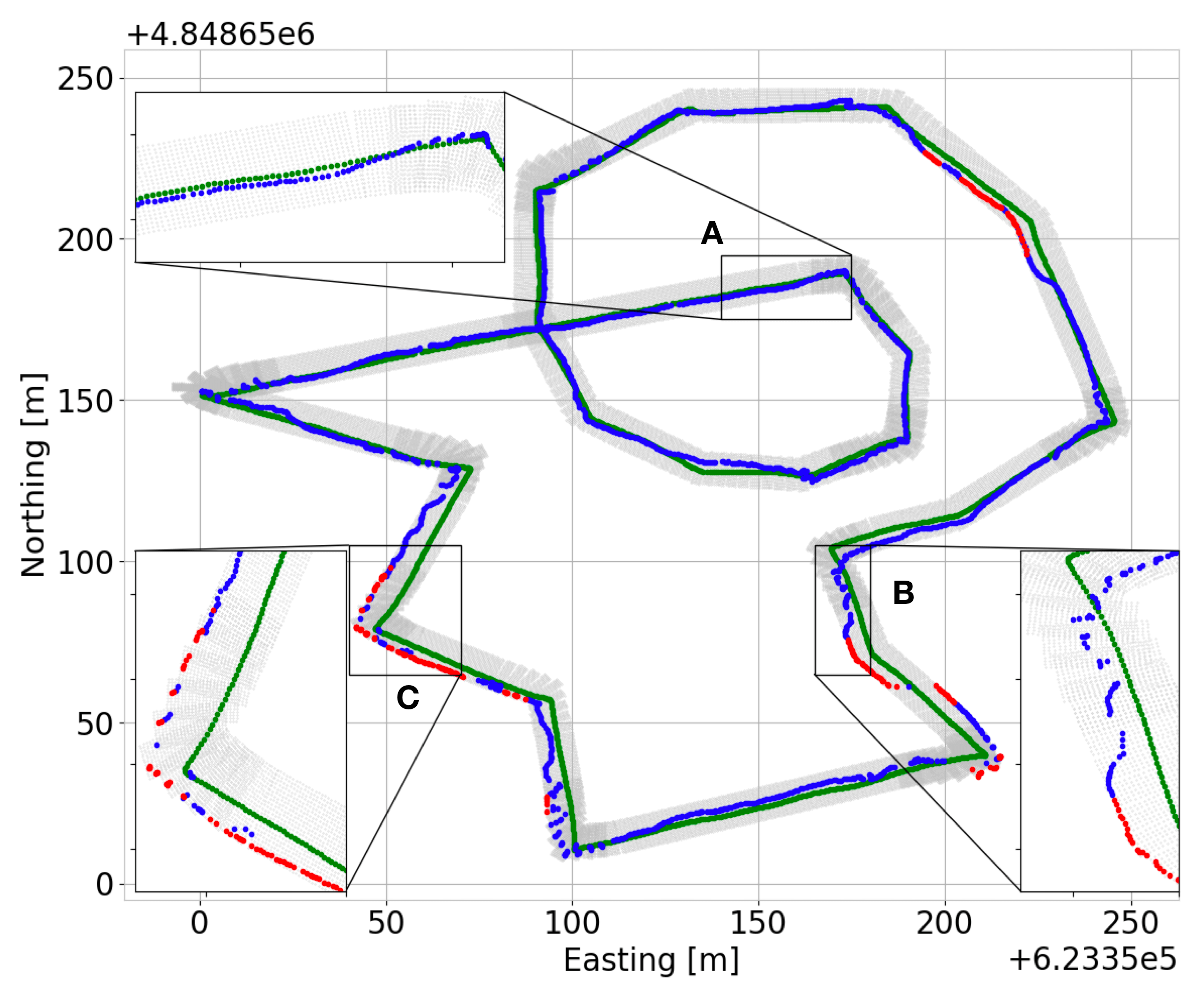}%
  \label{fig:evening}
}
\caption{Registration results showing our best (morning) and worst (evening) localization results. \change{Grey dots indicate the reference image positions. Green dots indicate the ground truth live image positions. Blue dots indicate accepted localizations and red dots indicate rejected localizations.} Shadows on the opposite sides of objects as compared to the GE reference images cause higher errors and more rejected registrations in the evening run.}
\label{fig:topdown}
\end{figure} 

\begin{figure*}
    \centering
    \includegraphics[width=\textwidth]{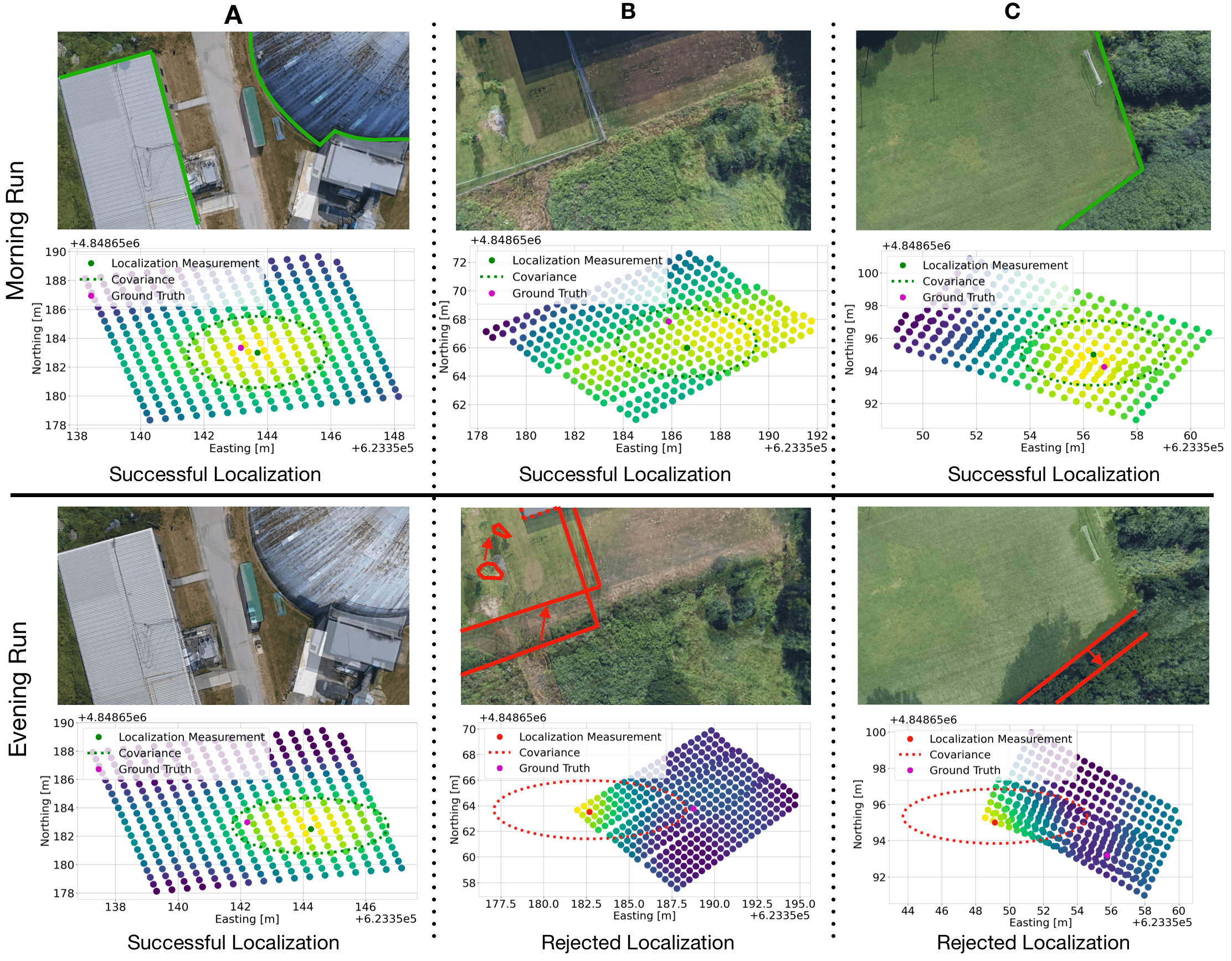}
    \caption{An example frame from each of the three areas indicated in Figure \ref{fig:topdown} for the morning lighting condition is shown at the top of this figure and for the evening lighting condition on the bottom. For each lighting condition, the top row shows the overlay between the live image and the GE reference image closest to the localization. The heat maps plotted in the bottom row show the value of the weights for each of the nearby GE reference images with yellow indicating a higher weight. The resulting localization and covariance is shown in green for successful registrations and in red for rejected registrations. The ground truth is shown in magenta. }
    \label{fig:testpoints}
    \vspace{-3mm}
\end{figure*}

In comparison with \cite{Patel2020}, for all but the noon lighting condition we achieve lower RMSE error on the longitude coordinate. For the latitude coordinate, we have lower RMSE error for three of the runs and for the other three runs we are an average of 0.3m higher. Our success rate of registrations is higher for all the lighting conditions. Particularly in the evening and sunset runs, we see an increase of \textasciitilde 10\% in the success rate and a decrease in RMSE.

The most significant advantage of our method over \cite{Patel2020} is the substantial reduction in runtime. Both methods were run on a Lenovo P52 laptop with an Intel i7 8th generation core, a Nvidia Quadro P2000 GPU, and 32 GB of RAM. Most of the MI registrations took between 5 to 35 seconds per frame, whereas our approach only took between 0.09 and 0.15 seconds. We were able to eliminate the costly optimization component from \cite{Patel2020}, which requires warping the image and recomputing the MI up to 150 times per registration. Instead, by rendering more reference images at a finer grid spacing and using the mean of the kernel weights to interpolate between them, we were able to achieve similar results at greatly reduced computation time.  

Both approaches are still limited by the storage available on the UAV. By autoencoding the reference images, we only need 1000 numbers to represent each image. Recording these numbers as half precision floats only requires 4.2 kB per image. In \cite{Patel2020}, the reference images are stored as $560\times315$ \mbox{4-bit} greyscale images requiring approximately 11 kB of storage which is almost three times as large. As a result of this reduction, we are able to render more images per meter of the path while still having a lower per meter storage cost, 0.241 Mb compared to 0.722 Mb. Encoding the images also makes the base comparison computation faster. An inner-product computation between two 1000 dim vectors takes on average 0.26ms, whereas a MI computation between two 176,400 pixel images takes on average 109ms. The disadvantage of our approach is that in addition to the autoencoded reference images, we must also store the weights for the trained neural networks in order to encode the live image on board. However, this is a fixed cost that does not increase with the length of the path. So for the 1.1km path in the dataset used, our total storage is still less than what is used in \cite{Patel2020}. Computation and storage comparisons are summarized in Table \ref{table:costs}.

We plot our registrations on the run with our best localization results, morning, in Figure \ref{fig:morning} and on the run with the worst localization results, evening, in Figure \ref{fig:evening}. The ground truth from the RTK is shown by the green dots. Successful registrations are indicated by blue dots and rejected registrations are indicated by red dots. The grey dots show the positions of the GE reference images. The localizations on the morning lighting condition likely perform the best because the shadow conditions match those on the GE images. In the evening dataset, the shadows are on the opposite sides of the objects as compared to the GE images. Figure \ref{fig:lc_imgs} shows an example image from each of the six lighting conditions for comparison.

\begin{table}[h]
\renewcommand{\arraystretch}{1.2}
\vspace{4mm}
\caption{Runtime and Storage Requirements Comparison}
\begin{tabular} {m{3.2cm}||m{1.9cm}|m{1.9cm}|| }
 \thead{\textbf{Comparison} \textbf{Method}} & \thead{Kernel (Ours)} & \thead{MI \cite{Patel2020}} \\
 \hline
 \makecell{Average Runtime \\ for 1.1 km Path}  &   \makecell[c]{\textbf{221 s}} &  \makecell[c]{18422 s} \\ 
 \hline
 \makecell{Average Time per Frame} &  \makecell[c]{\textbf{0.11 s}} &   \makecell[c]{9.23 s}\\ 
 \hline
\makecell{ Average Time per \\Comparison Computation }    & \makecell[c]{\textbf{0.26 ms}} & \makecell[c]{109 ms} \\ 
\hline
\makecell{Storage Cost per Image} & \makecell[c]{\textbf{4.2 kB}} &      \makecell[c]{11 kB}\\ 
\hline
\makecell{Storage Cost per m of Path} & \makecell[c]{\textbf{0.241 Mb}} &      \makecell[c]{0.722 Mb}\\ 
\hline
\makecell{Fixed Storage Cost}  &  \makecell[c]{158 Mb} & \makecell[c]{\textbf{0 Mb}}  \\ 
\hline
\makecell{Total Cost for 1.1km Path} & \makecell[c]{\textbf{423 Mb}}   & \makecell[c]{794 Mb}
\end{tabular}
\vspace{-6mm}
\label{table:costs}
\end{table}

We show an example from three different areas along the path for the morning and evening lighting conditions in Figure \ref{fig:testpoints}. The top row for each lighting condition shows an overlay of the live image and the GE reference image closest to the localization. The bottom row shows a heat map of the value of the weights of the nearby GE reference images computed from an inner-product with the live image. The resulting localization from the mean of the thresholded weights and the covariance estimate are plotted as well. For successful registrations, the covariance envelope is smaller. For the rejected localizations (i.e., evening B and C), shadows cause the highest weights to occur on misaligned images at the very edge of the reference images. In these cases, the covariance that results from the weights is larger than in the successful registrations, very elongated, and does not pass our threshold for outlier rejection. 

\section{CONCLUSIONS AND FUTURE WORK}

We presented a method for localizing live images captured from a UAV under six different lighting conditions to prerendered images from GE. Compared to the best existing method, we are able to achieve a similar level of accuracy at 1\% of the computation time. Our method also has a lower storage requirement per length of path making it an ideal candidate for running on board the UAV in future work. All preprocessing can be completed offline and grants the UAV the ability to traverse new areas without having to manually fly and map them first. Since our method is able to match images across large periods of time (the GE images are at least two years older), it could also be used for repeated traversals of the same path over large periods of time. Future work aims at integrating our method into a filtering pipeline such that it can be used in the loop on board the UAV.


\section*{ACKNOWLEDGMENT}
This work was funded by NSERC Canada Graduate Scholarship-Master's, Defence Research and Development Canada, Drone Delivery Canada, the Centre for Aerial Robotics Research and Education, University of Toronto, and the Vector Scholarship in Artificial Intelligence.

\bibliography{references}

\end{document}